\documentclass[runningheads]{llncs}
\usepackage{graphicx}
\usepackage{todonotes}
\usepackage{amsmath}
\usepackage{amssymb}
\usepackage{multirow}
\usepackage{booktabs}
\usepackage{wrapfig,lipsum,booktabs}

\newcommand{\otoprule}{\midrule[\heavyrulewidth]}
\newcommand{\printfnsymbol}[1]{%
  \textsuperscript{\@*}%
  }
\usepackage[ruled]{algorithm2e}

\begin{document}
\title{BrainTorrent: A Peer-to-Peer Environment for Decentralized Federated Learning}
\titlerunning{\emph{BrainTorrent}}
%
\author{Abhijit Guha Roy\inst{1,2}\thanks{The authors contributed equally.} \and Shayan Siddiqui\inst{1,2}\printfnsymbol{1} \and
Sebastian P\"olsterl\inst{1} \and \\
Nassir Navab\inst{2}\and
Christian Wachinger\inst{1}
}
\authorrunning{Roy et al.}
%
\institute{Artificial Intelligence in Medical Imaging (AI-Med), KJP, LMU Munich, Germany
\and
Computer Aided Medical Procedures, Technische Universit\"{a}t M\"{u}nchen, Germany}
\maketitle              
\begin{abstract}
Access to sufficient annotated data is a common challenge in training deep neural networks on medical images. As annotating data is expensive and time-consuming, it is difficult for an individual medical center to reach large enough sample sizes to build their own, personalized models. As an alternative, data from all centers could be pooled to train a centralized model that everyone can use. However, such a strategy is often infeasible due to the privacy-sensitive nature of medical data. Recently, federated learning (FL) has been introduced to collaboratively learn a shared prediction model across centers without the need for sharing data. In FL, clients are locally training models on site-specific datasets for a few epochs and then sharing their model weights with a central server, which orchestrates the overall training process. Importantly, the sharing of models does not compromise patient privacy. A disadvantage of FL is the dependence on a central server, which requires all clients to agree on one trusted central body, and whose failure would disrupt the training process of all clients. In this paper, we introduce \emph{BrainTorrent}, a new FL framework without a central server, particularly targeted towards medical applications. BrainTorrent presents a highly dynamic peer-to-peer environment, where all centers directly interact with each other without depending on a central body. We demonstrate the overall effectiveness of FL for the challenging task of whole brain segmentation and observe that the proposed server-less BrainTorrent approach does not only outperform the traditional server-based one but reaches a similar performance to a model trained on pooled data.

\end{abstract}
\section{Introduction}
\label{sec:intro}
Training deep neural networks (DNNs) effectively requires access to abundant annotated data. 
This is a common concern in medical applications, where annotations are both, expensive and time-consuming.
For instance, manual labeling of a single 3D brain MRI scan can take up to a week by a trained neuroanatomist~\cite{fischl2002whole}. 
It is therefore challenging to reach sample sizes with in-house curated datasets that enable an effective application of deep learning. 
Pooling data across medical centers could alleviate the limited data problem. 
However, data sharing is restricted due to ethical and legal regulations, preventing the aggregation of medical data from multiple medical centers. 
This leaves us with a scenario, where each center might have too limited data to effectively train DNNs but the combination of data across the centers that work on the same problem is not possible. 

In such a decentralized environment, where  data is distributed across  centers, training a common DNN is challenging. One na\"ive approach would be to train in a sequentially incremental fashion~\cite{sheller2018multi}. Such an approach trains the network at a given center and then passes the DNN weights to the next center, where it is fine-tuned on new data, and so on. A common problem encountered with such an approach is catastrophic forgetting \cite{french1999catastrophic}, i.e., at every stage of fine-tuning, the new data overwrites the knowledge acquired from the previous training, thus deteriorating generalizability. Also, certain centers can have very limited data, which can risk overfitting the network severely.

Recently, a learning strategy was introduced to address challenges in training DNNs in such decentralized environments called \emph{Federated Learning} (FL)~\cite{mcmahan2017}. The environment consists of a central server, which is connected to all the centers coordinating the overall process. The main motivation of this framework is to provide assistance to mobile device users. In such a scenario, where users can easily scale into the millions, it is very convenient to have a central server body. Also, one of the main aims of FL is to minimize communication costs. In the scenario of collaborative learning within a community, like medical centers, the motivations are a bit different. Firstly, in contrast to millions of clients in FL, the number of medical centers forming a community is much lower (in the order of 10s). Secondly, each center can be expected to have a strong communication infrastructure, so that communication cycles are not a big bottleneck. Thirdly, it is difficult to have a central trusted server body in such a setting, rather every center may want to coordinate with the rest directly. Fourthly, if the whole community is dependent on the server and a fault occurs at the server, the whole system is non-operational, which is undesirable in a medical setting.

In this paper, we propose for the first time a server-less, peer-to-peer approach to federated learning where clients communicate directly among themselves.  We term this decentralized environment \textit{The BrainTorrent}. The design is motivated to fulfill the above mentioned requirements for a group of medical centers to collaborate. Absence of a central server body not only makes our environment resistant to failure but also precludes the need for a body everyone trusts. 
Further, any client at any point can initiate an update process very dynamically. 
As number of communication round is not a bottleneck for medical centers, they can interact more frequently.
Due to this high frequency of interaction and update process per client, the models in \emph{BrainTorrent} converges faster and reaches an accuracy similar to a model trained with pooling the data from all the clients.   

The main contributions of the paper: (i) we introduce \emph{BrainTorrent} a peer-to-peer, decentralized environment where multiple medical centers can collaborate and benefit from each other without sharing data among them, (ii) we propose a new training strategy of DNNs within this environment in a federated learning fashion that does not rely upon a central server-body to coordinate the process, and (iii) we demonstrate exemplar cases where different centers in the environment can have data with different age ranges and non-uniform data distribution, where \emph{BrainTorrent} outperforms traditional, server-based federated learning.

\noindent
\textbf{Prior work: } Federated learning (FL) was proposed by Mcmahan et. al.~\cite{mcmahan2017} for training models from decentralized data distributed across different clients (mainly mobile users), preserving data privacy. Other works build on top of FL by improving communication efficiency~\cite{konevcny2016federated}, improving system scalability~\cite{bonawitz2019towards} and improving encryption for better privacy~\cite{bonawitz2017practical}. The traditional FL concept has recently been applied to medical image analysis for 2-class segmentation of brain tumors~\cite{sheller2018multi}, demonstrating its feasibility. In contrast to that, we introduce a new peer-to-peer FL approach, and tackle the more challenging task of whole-brain segmentation with $20$ classes with severe class-imbalance. To the best of our knowledge, this is the first application of FL to whole-brain segmentation.

\section{Method} 
\label{sec:method}
Let us consider an environment with $N$ centers $\{ C_1, \dots, C_N \}$, where each center $C_i$ has training data $\mathcal{D}_i = \{ (x_1, y_1), \dots, (x_{a_i}, y_{a_i}) \}$ with $a_i$ labeled samples. 
In a general setting,  data from all centers are pooled together ($a = \sum_i a_i$) at a common server, where a common model is learned that is distributed across all the centers for usage. 
In a medical setting, data in each center is very sensitive containing patient specific information, which cannot be shared across  centers or with a central server $\mathcal{S}$.

\begin{figure*}[t]
\center{\includegraphics[width=0.95\textwidth]{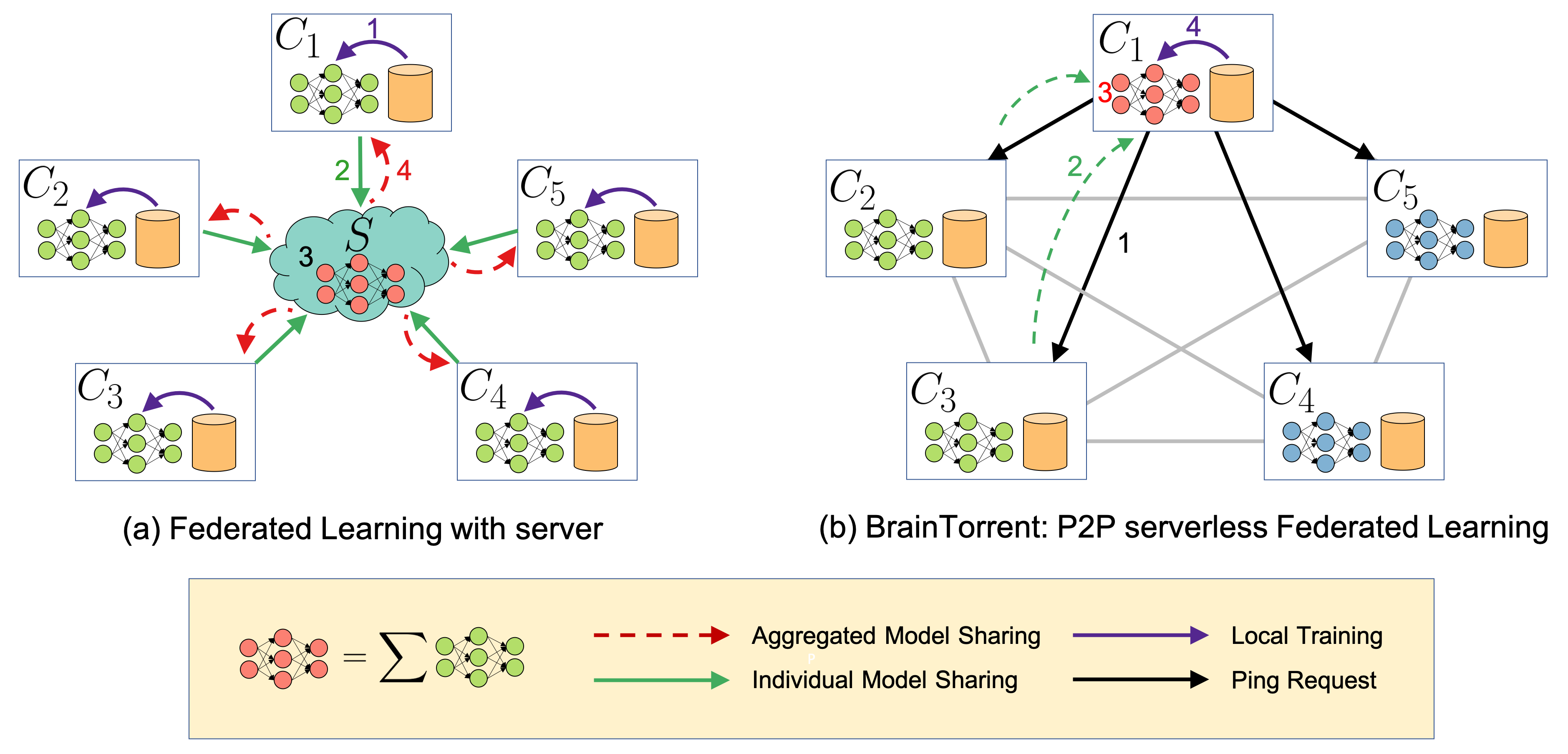}}
\caption{Overview of the federated learning (FL) process with $5$ clients. FL with server (FLS) is shown in (a), and our proposed peer-to-peer (P2P) server-less \emph{BrainTorrent} in (b). A single round of training is shown for both that consists of 4 steps. In FLS: 1. Each client updates (updated model in green) on local data (purple line), 2. All clients send their models to the server ($\mathcal{S}$) (green line), 3. Server aggregates all models (in red) and, 4. The aggregated model is sent back to clients (red dotted lines).
In \emph{BrainTorrent}: 1. A client ($C_1$) sends a ping request (black lines) to all other clients for checking their versions, 2. Clients $C_2$ and $C_3$ have a newer version (in green) whereas $C_4$ and $C_5$ do not (in blue). Weights of only $C_2$ and $C_3$ are sent to $C_1$ (green dotted lines), 3. An aggregated model (in red) is formed combining $C_1$, $C_2$ and $C_3$, 4. The aggregated model is fine-tuned on local data of $C_1$ (purple line).}
\label{fig:FL_illustration}
\end{figure*}

\subsection{Federated Learning with Server}
The process of distributed training in traditional FL with server (FLS) is illustrated in Fig.~\ref{fig:FL_illustration} (a). Here, in a single round of training, all $N$ clients $\{ C_i \}_{i=1}^N$ start with training their respective networks in parallel only for a few iterations (not till convergence). 
Let the weight parameter list for all clients be indicated by $\{ \mathbf{W}^1, \dots, \mathbf{W}^N \}$.
Next, all the clients send these partially trained parameters to the central server $\mathcal{S}$, which aggregates them by weighted averaging $\mathbf{W}^{\mathcal{S}} = \sum_i \frac{a_i}{a} \mathbf{W}^i$. The multiplicative factor is computed as the fraction of the total data belonging to a client. 
The rationale is to emphasize clients with more training data.
Finally, the aggregated model $\mathbf{W}^{\mathcal{S}}$ is distributed back to all clients for further training. We refer to~\cite{mcmahan2017} for a more detailed description of the implementation. 

Several rounds are executed until all client models converge. At the end of the training process, each client has its own personalized model $\mathbf{W}^i$, fine-tuned to its local data, and the server model $\mathbf{W}^{\mathcal{S}}$, which is more generic to new unseen data. The central server $\mathcal{S}$ has the vital role of coordinating the aggregation and re-distributing the weights across clients. When a new client is added to the environment, it receives the server model $\mathbf{W}^{\mathcal{S}}$ to start off.

\subsection{\emph{BrainTorrent}: Server-less Peer-to-Peer Federated Learning}
We introduce \emph{BrainTorrent}, a peer-to-peer FL environment within a community-based environment, illustrated in Fig.~\ref{fig:FL_illustration} (b). 
As there is no central server for coordinating the training process, a new strategy is required, as explained in the following. 
Firstly, all clients $\{C_i \}_{i=1}^N$ in this environment are connected directly in a peer-to-peer fashion, as indicated in Fig.~\ref{fig:FL_illustration} (b). Unlike FLS, along with the model, each client maintains a vector $\mathbf{v} \in \mathbb{N}^N$ containing its own version and the last versions of models it used during merging. At the start, every entry is initialized to zero. Every time a fine-tuning step occurs, it increments its own version number. The training process at any step is conducted with the following steps:
\begin{enumerate}
    \item Locally train each client in parallel for a few iterations using local dataset.
    \item A random client $C_i$ from the environment initiates the training process. It sends out a `ping\_request' to the rest of the clients to get their latest model versions to generate $\mathbf{v}_{\text{new}}$. $\mathbf{v}_{\text{old}}$ is initiated with client's $\mathbf{v}$.
    \item All clients $C_j$ with updates, i.e., $v^j_{\text{old}} < v^j_{\text{new}}$, send their weights $\mathbf{W}^j$ and the training sample size $a_j$ to $C_i$.
    \item This subset of models is merged with $C_i$'s current model to a single model by weighted averaging. Then return to Step 1.
\end{enumerate}

This comprises a single round of training. It must be noted that the definition of a round here is different from FLS. In a single round of FLS, all clients are updated once by fine-tuning, whereas in our framework only a single random client is updated. So, the number of updates per client in $R$ rounds of FLS is equivalent to $R \times N$ rounds of our framework.
The steps are presented in Algorithm~\ref{algo:decentralized}.

\begin{figure}[t]
\begin{algorithm}[H]
 \caption{Decentralized training of \emph{BrainTorrent}. }
 Initialize $N$ clients models, $\mathbf{C}=\{C_1, \dots , C_N \}$ with random weights\;
 Initialize $N$ version vectors $\mathbf{V}=[\mathbf{v}^1, \dots, \mathbf{v}^N]$ with all zero entries\;
 \For {round r in 1, 2, \dots}{
  Randomly select a client $i$ from \{$1,\ldots,N\}$\;
  $\mathbf{v}_{\mathrm{old}} \leftarrow \mathbf{v}^i$\;
  $\mathbf{v}_{\mathrm{new}} \leftarrow$ ping\_request($C_i \rightarrow \mathbf{C}$)\;
  $\mathbf{W} \leftarrow \frac{a_i}{a}\mathbf{W}^i$ \;
  \For{$j \in \{1, \dots, i-1, i+1, \ldots, N\}$}{
  \If{$v^j_{\mathrm{new}} > v^j_{\mathrm{old}}$}{
    Receive updated $\mathbf{W}^j$ and $a_j$ from $C_j$ \;
   }
   $\mathbf{W} \leftarrow \mathbf{W} + \frac{a_j}{a}\mathbf{W}^j$ \;
  }
  $\mathbf{W}^i \leftarrow$
  FineTune($\mathbf{W}$, $\mathcal{D}_i$) \;
  Increment $\mathbf{v}^i(i)$\; 
 }
 \label{algo:decentralized}
\end{algorithm}
\end{figure}

\section{Experimental Settings}
To demonstrate the effectiveness of BrainTorrent, we choose the challenging task of whole-brain segmentation of MRI T1 scans. 
We use the Multi-Atlas Labelling Challenge (MALC) dataset~\cite{landman2012miccai} for our experiments. The dataset consists of $30$ annotated whole-brain MRI T1 scans from different patients, out of which we always use $20$ scans for training and the remaining $10$ for testing. Manual annotations were provided by Neuromorphometrics Inc. As a segmentation network, we decided to use the QuickNAT architecture~\cite{roy2019quicknat}, which demonstrated state-of-the-art performance for whole-brain segmentation. We combined the left and right brain structures in one class and all the cortical parcellations in a single cortex class, thus reducing the task to a 20-class segmentation problem. During fine-tuning at each client center, we fix the number of epochs to $2$, without risking any client-specific overfitting. Initially, all clients had a learning rate of $0.001$, which was reduced by a factor of $0.5$ after every $4$ update rounds. We use Adam for optimization. We explore two experimental settings detailed below, where we compare our proposed \emph{BrainTorrent} and FLS.



\subsection{Experiment 1}
In this experiment, we randomly distributed the $20$ training scans among the clients in a uniform fashion, i.e., each client receives the same number of scans. Here, we also conduct experiments by varying the number of clients by $\{ 5, 7, 10, 20 \}$ in the environment.  

\paragraph{\textbf{Increase in Number of Clients}}
We investigate the FL performance as the number of clients increases in the environment. Since the number of training scans are fixed to $20$, scans per client reduces with increasing clients. We vary the number of clients between $\{ 5, 7, 10, 20 \}$, which results in $\{ 4, 3, 2, 1 \}$ training scans per client, respectively. For the setting with $7$ clients, one client had $2$ scans whereas the rest had $3$ scans.
Under this setting, we compare FL and BrainTorrent and present the results in Tab.~\ref{tab:num_clients}. We compare the average Dice score across all clients for FLS and BrainTorrent for all the configurations on the $10$ test scans. Also, we compare their aggregated model, i.e., the model which would be provided to a new client when it first joins the environment. For FL, this is the server model, whereas for BrainTorrent, we create a model by averaging the model weights of all the clients in the environment.
As an upper bound model, we trained a model by pooling all the data across the clients termed as `pooled model'.
We observe that irrespective of the number of clients,  BrainTorrent outperforms FLS for both, average Dice score over clients and Dice score for aggregated model. Also, as the number of clients increases (and therefore number of scans per client decreases), the performance degrades. This drop in performance is only marginal at the beginning up to $10$ clients, and drops by a huge margin when each center has only $1$ annotated scan, simulating an extremely limited data scenario. For the aggregated model, we observe that BrainTorrent outperforms FL by $1-2\%$ Dice points. Also, we observe that BrainTorrent achieves the same level of segmentation accuracy that would be reached by the `pooled model', which is striking given the constraints. This performance is sustained for number of clients 5 to 7 with only 4-3 training scans per center. 

\begin{table*}[h]
\centering
\scriptsize
\caption{Experiment showing variation of performance with changing number of Clients in the environment for both FLS and \emph{BrainTorrent} along with `Pooled Model'.}
\begin{tabular}{llcccc}
\toprule
\multirow{2}{*}{\# Clients}& \multirow{2}{*}{Scans/client} & \multicolumn{2}{c }{Avg. Dice over Clients} & \multicolumn{2}{c }{Aggregated Model} \\ \cmidrule(r){3-4} \cmidrule(r){5-6}
& & FLS & BrainTorrent & FLS & BrainTorrent \\
 \otoprule
5 & 4 & $0.812$ & $0.851$ & $0.845$ & $\mathbf{0.863}$    \\
7 & 3 & $0.753$ & $0.837$ & $0.843$ & $0.861$ \\
10 & 2 & $0.792$ & $0.807$ & $0.842$ & $0.850$   \\
20 & 1 & $0.570$ & $0.578$ & $0.687$ & $0.728$   \\ \hline
\multicolumn{2}{c }{Pooled  Model} & \multicolumn{4}{c}{$\mathbf{0.866}$} \\
\bottomrule 
\end{tabular}
\label{tab:num_clients}
\end{table*}

\paragraph{\textbf{Analysis of client-wise performance}}
We take a closer look at the segmentation performance of the client-specific models. We select the configuration with $10$ clients, where every client has $2$ annotated training scans. We report the performance of each client model for both BrainTorrent and FLS in Tab.~\ref{tab:per_clients_analysis}. Also, as lower bound analysis, we train client-specific models with only $2$ scans, referred to as `only client' models and report their performance on the same validation set.
First, we observe that both FLS and BrainTorrent outperform the `only client' model by an average of $24\%$ and $26\%$ Dice points, substantiating the immense effectiveness of the federated learning approach. Further, BrainTorrent achieves an average $2\%$ higher Dice score than FLS, where $7$ out of $10$ client models performed better in BrainTorrent than in FLS. 
This reaffirms our claim that BrainTorrent does not only result in a stronger aggregated model but also in more robust client-level personalized models.

\begin{table*}[t]
\centering
\scriptsize
\caption{Per-client performance analysis comparing FLS and \emph{BrainTorrent} for $10$ clients, each with $2$ training scans. This is compared against model trained exclusively on client data (Only Client) without FL.}
\begin{tabular}{lp{0.8cm}p{0.8cm}p{0.8cm}p{0.8cm}p{0.8cm}p{0.8cm}p{0.8cm}p{0.8cm}p{0.8cm}p{0.8cm}p{0.8cm}}
\toprule
Method & $C_1$ & $C_2$ & $C_3$ & $C_4$ & $C_5$ & $C_6$ & $C_7$ & $C_8$ & $C_9$ & $C_{10}$ & Mean \\
 \otoprule
BrainTorrent & $0.806$ & $0.781$ & $0.835$ & $0.818$ & $0.793$ & $0.819$ & $0.812$ & $0.771$ & $0.820$ & $0.817$ & $\mathbf{0.807}$ \\
FLS & $0.808$ & $0.794$ & $0.779$ & $0.789$ & $0.784$ & $0.770$ & $0.790$ & $0.784$ & $0.809$ & $0.814$ & $0.792$ \\ \hline
Only Client & $0.570$ & $0.635$ & $0.578$ & $0.609$ & $0.526$ & $0.502$ & $0.603$ & $0.570$ & $0.513$ & $0.510$ & $0.564$ \\
\bottomrule 
\end{tabular}
\label{tab:per_clients_analysis}
\end{table*}

\vspace{-0.3cm}
\subsection{Experiment 2}
\vspace{-0.2cm}
In this experiment, we distribute the $20$ training scans across $5$ clients, where each client has scans for a specific, non-overlapping age range, see Tab.~\ref{tab:experiment2}.
This experiment simulates the scenario that each client has data with unique characteristics.
In addition, it also provides a scenario for non-uniform data distribution, where the number of training scans differs among clients, yielding a realistic clinical use-case.

\begin{wraptable}{r}{6cm}
\centering
\scriptsize
\vspace{-1cm}
\caption{Data distribution for Exp 2}
\begin{tabular}{p{1.5cm}cc}
\toprule
Clients & Age Range & Training scans \\ 
 \otoprule
Client 1 & $\leq 20$ & $5$ vols \\
Client 2 & $(20; 30]$ & $9$ vols \\
Client 3 & $(30; 40]$ & $2$ vols \\
Client 4 & $(40; 50]$ & $1$ vols \\
Client 5 & $\geq 50$ & $3$ vols \\
\bottomrule 
\end{tabular}
\label{tab:experiment2}
\vspace{-0.5cm}
\end{wraptable}

Tab.~\ref{tab:age_specific_clients} reports the results for FLS and BrainTorrent. 
We observe that under such an uneven distribution, the aggregated model of BrainTorrent achieves the performance of `pooled model', whereas FLS had a performance $3\%$ Dice points below that.
Comparing average Dice scores across clients, BrainTorrent outperforms FLS by a margin of $7\%$ Dice points.
This demonstrates that in a scenario of non-uniform data distribution, performance of BrainTorrent is unaffected, whereas FLS performance degrades.
Also, it must be noted that the performance of $C_3$ and $C_4$, which have only 2 and 1 annotated scans, respectively, is comparatively low for FLS. One possible cause can be slight overfitting. In contrast, these clients perform very well in the BrainTorrent framework.

\section{Conclusion}
In this paper, we introduced \emph{BrainTorrent}, a server-less peer-to-peer federated learning environment for decentralized training. In contrast to traditional FL with server, our framework does not rely on a central server body for orchestrating the training process. We presented a proof-of-concept study tackling the challenging task of whole-brain segmentation, training a complex fully convolutional neural network in a decentralized fashion. We demonstrated in our experiments that BrainTorrent achieves a better performance than FLS under different experimental settings. The margin extends up to $7\%$ Dice points in scenarios where clients have unequal numbers of training scans.
Overall, BrainTorrent does not only resolve the issue relying on a central server but also enables more robust training of clients through highly dynamic updates, reaching performance similar to a model trained on data pooled across clients.
Although we focused on image segmentation, our proposed method is generic and can be used for training any machine learning model.

\begin{table*}[t]
\centering
\scriptsize
\caption{Comparison of BrainTorrent and FLs for Exp. 2 as detailed in Tab.~\ref{tab:experiment2}}
\begin{tabular}{p{2cm}ccccccc}
\toprule
Method & \multicolumn{5}{c}{Client-wise Dice} & Avg. Dice & Aggregated \\ \cmidrule(r){2-6} 
& $C_1$ & $C_2$ & $C_3$ & $C_4$ & $C_5$  & per client & Model \\
 \otoprule
BrainTorrent & $0.853$ & $0.852$ & $0.856$ & $0.839$ & $0.857$ & $\mathbf{0.851}$ & $\mathbf{0.864}$ \\
FLS & $0.807$ & $0.804$ & $0.729$ & $0.731$ & $0.792$ & $0.772$ & $0.828$ \\
 \hline
Pooled Model & \multicolumn{7}{c}{$\mathbf{0.866}$}\\
\bottomrule 
\end{tabular}
\label{tab:age_specific_clients}
\end{table*}

%
%
%
%
\bibliographystyle{splncs04}
\bibliography{references.bib}

\begin{thebibliography}{1}
\providecommand{\url}[1]{\texttt{#1}}
\providecommand{\urlprefix}{URL }
\providecommand{\doi}[1]{https://doi.org/#1}

\bibitem{bonawitz2019towards}
Bonawitz, K., Eichner, H., Grieskamp, W., et~al.: Towards federated learning at
  scale: System design. arXiv preprint arXiv:1902.01046  (2019)

\bibitem{bonawitz2017practical}
Bonawitz, K., Ivanov, V., Kreuter, B., Marcedone, A., McMahan, H.B., Patel, S.,
  Ramage, D., Segal, A., Seth, K.: Practical secure aggregation for
  privacy-preserving machine learning. In: Proceedings of the 2017 ACM SIGSAC
  Conference on Computer and Communications Security. pp. 1175--1191. ACM
  (2017)

\bibitem{fischl2002whole}
Fischl, B., Salat, D.H., Busa, E., Albert, M., Dieterich, M., Haselgrove, C.,
  Van Der~Kouwe, A., Killiany, R., Kennedy, D., Klaveness, S., et~al.: Whole
  brain segmentation: automated labeling of neuroanatomical structures in the
  human brain. Neuron  \textbf{33}(3),  341--355 (2002)

\bibitem{french1999catastrophic}
French, R.M.: Catastrophic forgetting in connectionist networks. Trends in
  cognitive sciences  \textbf{3}(4),  128--135 (1999)

\bibitem{konevcny2016federated}
Kone{\v{c}}n{\`y}, J., McMahan, H.B., et~al.: Federated learning: Strategies
  for improving communication efficiency. arXiv preprint arXiv:1610.05492
  (2016)

\bibitem{landman2012miccai}
Landman, B., Warfield, S.: Miccai 2012 workshop on multi-atlas labeling. In:
  MICCAI (2012)

\bibitem{mcmahan2017}
McMahan, B., Moore, E., Ramage, D., Hampson, S., y~Arcas, B.A.:
  Communication-efficient learning of deep networks from decentralized data.
  In: AISTATS. pp. 1273--1282 (2017)

\bibitem{roy2019quicknat}
Roy, A.G., Conjeti, S., Navab, N., Wachinger, C.: Quicknat: A fully
  convolutional network for quick and accurate segmentation of neuroanatomy.
  NeuroImage  \textbf{186},  713--727 (2019)

\bibitem{sheller2018multi}
Sheller, M.J., Reina, G.A., Edwards, B., Martin, J., Bakas, S.:
  Multi-institutional deep learning modeling without sharing patient data: A
  feasibility study on brain tumor segmentation. In: MICCAI Brainlesion
  Workshop. pp. 92--104. Springer (2018)

\end{thebibliography}

\end{document}